\pdfoutput=1

\documentclass[11pt]{article}

\usepackage{acl}

\usepackage{times}
\usepackage{latexsym}

\usepackage[T1]{fontenc}

\usepackage[utf8]{inputenc}

\usepackage{microtype}

\usepackage{CJKutf8}
\usepackage{tabularx}

\title{HuaTuo (\begin{CJK*}{UTF8}{gbsn}华驼\end{CJK*}): Tuning LLaMA Model with Chinese Medical Knowledge}

 \author{
 Haochun Wang$^{*}$, Chi Liu$^{*}$, Nuwa Xi, Zewen Qiang, Sendong Zhao,  Bing Qin and Ting Liu \\
 Research Center for Social Computing and Information Retrieval, \\Harbin Institute of Technology, China \\
 \{hcwang,cliu,nwxi,zwqiang,sdzhao,bqin,tliu\}@ir.hit.edu.cn
 }

\begin{document}
\maketitle

\renewcommand{\thefootnote}{\fnsymbol{footnote}}
\footnotetext[1]{Equal contribution.}
\renewcommand{\thefootnote}{\arabic{footnote}}

\begin{abstract}
Large Language Models (LLMs), such as the LLaMA model, have demonstrated their effectiveness in various general-domain natural language processing (NLP) tasks. Nevertheless, LLMs have not yet performed optimally in biomedical domain tasks due to the need for medical expertise in the responses. In response to this challenge, we propose HuaTuo (\begin{CJK*}{UTF8}{gbsn}华驼\end{CJK*}), a LLaMA-based model that has been supervised-fine-tuned with generated QA (Question-Answer) instances. The experimental results demonstrate that HuaTuo generates responses that possess more reliable medical knowledge. Our proposed HuaTuo model is accessible at \url{https://github.com/SCIR-HI/Huatuo-Llama-Med-Chinese}.
\end{abstract}
\section{Introduction}

The advent of instruction-following large language models (LLMs), representative by ChatGPT\cite{chatgpt}, has generated significant interest due to their exceptional performance in understanding instructions and generating human-like responses. Compared to smaller models, LLMs exhibit strong generalization across various natural language processing (NLP) tasks and unique emergent ability to solving unseen or complicated tasks. Despite ChatGPT's non-open source status, open-source communities have provided several alternatives, such as LLaMa\citep{touvron2023llama}, with relatively affordable training costs. This positions LLMs as potential solutions for real-world scenarios requiring communication and reasoning.

However, despite their numerous merits, LLMs are not designed to cater specifically to the medical domain. Their general domain knowledge often falls short when addressing such specialized fields, where accurate and domain-specific expert knowledge is critical. This can lead to sub-optimal diagnostic precision, drug recommendations, and medical advice, potentially endangering patients. Few efforts have been made to address this problem, with existing approaches primarily focusing on supplying LLMs with medical information retrieved from conversations, where human errors may occur more frequently. Additionally, LLMs are typically trained in English, constraining their comprehension and response capabilities in languages that differ significantly from English, such as Chinese, rendering their direct application in Chinese contexts less than ideal.

In this paper, we present the HuaTuo (\begin{CJK*}{UTF8}{gbsn}华驼\end{CJK*}) model, an LLM tailored for the biomedical domain, focusing on the Chinese language. By generating diverse instruction data based on medical knowledge from CMeKG, we emphasize ensuring the correctness of facts in the model's responses, which is vital in the biomedical domain. Through this process, we collect over 8,000 instruction data for supervised fine-tuning. Our model builds upon the open-source LLaMa-7B base model, integrates structured and unstructured medical knowledge from the Chinese medical knowledge graph (CMeKG), and employs knowledge-based instruction data for fine-tuning.

In summary, our contributions can be summarized as follows: 

\begin{itemize}
    \item We introduce the HuaTuo model, the first open-source Chinese biomedical LLM tuned with knowledge-based instruction data;
    \item We integrate structured and unstructured medical knowledge from CMeKG, ensuring our model has accurate and domain-specific knowledge;
    \item We proposed SUS, a novel metric for evaluating LMs in the biomedical domain considering safety, usability and smoothness.
\end{itemize}

\section{Related Works}

\subsection{Large Language Models}
Recent advancements in large language models (LLMs) have demonstrated their superiority over previous-generation paradigms, such as pre-training and fine-tuning. The significant increase in model scale has led to qualitative changes in LLMs, commonly referred to as emergent abilities. These include in-context learning for zero-shot tasks and chains of thought that enhance the model's performance on complex tasks.

OpenAI's development of ChatGPT and GPT-4 has revolutionized the perception of LLMs. Although these models exhibit remarkable performance, OpenAI has not disclosed details regarding their training strategies or weight parameters. LLaMa serves as an open-source alternative for GPT, with sizes ranging from 7 billion to 65 billion parameters. \citeauthor{alpaca} trained Alpaca based on LLaMa with instruction tuning. 

While comparable in performance to GPT-3.5, LLaMa's performance on Chinese tasks is subpar due to its training data is primarily limited to English corpus. To address Chinese-specific applications, \citeauthor{du2022glm,zeng2023glm-130b} introduced GLM, a 130 billion-parameter auto-regressive pre-trained model with multiple training objectives. ChatGLM further incorporates code training and aligns with human intentions through supervised fine-tuning, offering a tailored solution for Chinese contexts.

\subsection{Pre-trained Models in Biomedical Domain}

Although large language models (LLMs) exhibit remarkable performance in general domains, their lack of domain-specific knowledge results in suboptimal performance in fields that require specialized expertise, such as biomedicine. The biomedical field's inherent nature necessitates models to possess comprehensive knowledge bases for relevant queries, particularly when applied to real-world situations where patients seek health and medical advice. Several efforts have been made to adapt LLMs to the biomedical domain.

Existing approaches primarily employ ChatGPT for data assistance and train smaller models using its distilled or translated knowledge. Chatdoctor\cite{yunxiang2023chatdoctor} represents the first attempt to adapt LLMs to the biomedical field by fine-tuning LLaMa using conversation demonstrations synthesized via ChatGPT. DoctorGLM\cite{xiong2023doctorglm} leverages ChatGLM-6B as the base model and fine-tunes it with the Chinese translation of ChatDoctor dataset, obtained through ChatGPT. Additionally, \citeauthor{llm-zoo-2023} develops a Chinese and Medically Enhanced Language model within their collection of LLMs. Collectively, these works illustrate the potential for LLMs to be successfully applied within the biomedical domain.



\section{HuaTuo Model}
In this section, we will introduce the training process of our HuaTuo (\begin{CJK*}{UTF8}{gbsn}华驼\end{CJK*}) model.

\subsection{Base Model}
LLaMA \cite{touvron2023llama} is a collection of multi-lingual base models with parameters ranging from 7 billion to 65 billion, which are open-source to the research community. Here, we adopted the LLaMA-7B model for more accessible training.

\subsection{Medical Knowledge}
There are various kinds of medical knowledge, generally including (1) structured medical knowledge like medical knowledge graphs and (2) unstructured medical knowledge like medical guidelines. We utilized a Chinese medical knowledge graph, CMeKG \cite{Odmaa}, which also provides retrieved medical knowledge about diseases, drugs, symptoms, etc. Table \ref{knowledge-label} shows several knowledge cases in the CMeKG knowledge base.
\begin{table*}[htbp]
\centering
\begin{tabularx}{\textwidth}{l|X|X}
\hline
\textbf{Type}    & \textbf{Knowledge in Chinese}   & \textbf{Knowledge translated to English} \\ \hline
Disease & \{\begin{CJK*}{UTF8}{gbsn}"class": "百种常见病", "中心词": "肝癌", "药物治疗": {[}"瑞格非尼", "对乙型或丙型肝炎有效的抗病毒药物", "索拉非尼"{]}, "多发地区": {[}"撒哈拉以南的非洲"{]}, "高危因素": {[}"肥胖", "HBV DNA过高", "慢性酗酒", "男性", "慢性乙型肝炎感染", "肝癌家族史", "慢性丙型肝炎肝硬化", "核心启动子突变", "肝硬化", "HCV重叠感染", "老年性心瓣膜病", "乙型肝炎e抗原", "糖尿病"{]},"发病部位": {[}"肝脏"{]}, "辅助检查": {[}"肝功能检查"{]}, "病史": {[}"长期慢性乙肝病史"{]}\end{CJK*}\} & \{"class": "Common Diseases", "Key Word": "Liver Cancer", "Drug Treatment": ["Regorafenib", "Antiviral drugs effective against hepatitis B or C", "Sorafenib"], "High Prevalence Regions": ["Sub-Saharan Africa"], "High Risk Factors": ["Obesity", "High HBV DNA levels", "Chronic alcoholism", "Male gender", "Chronic hepatitis B infection", "Family history of liver cancer", "Cirrhosis due to chronic hepatitis C", "Core promoter mutation", "Liver cirrhosis", "HCV co-infection", "Senile valvular heart disease", "Hepatitis B e antigen", "Diabetes"], "Affected Area": ["Liver"], "Auxiliary Examination": ["Liver function test"], "Medical History": ["Long-term history of chronic hepatitis B"]\}.                                           \\ \hline
Drug    &  \{\begin{CJK*}{UTF8}{gbsn} {"class": "西药", "中心词": "二甲双胍", "性状": ["糖衣或薄膜衣片，除去包衣后显白色"], "英文名称": ["异福片", "格华止"], "分类": ["双胍类", "抗结核病药"], "规格": ["0.25g"], "OTC类型": ["乙类OTC", "甲类OTC"], "适应证": ["糖尿病", "肥胖"], "通用名": ["异福片"], "成份": ["利福平及异烟肼", "异烟肼", "异烟肼0.1克", "异烟肼150毫克", "本品为复方制剂", "利福平", "利福平300毫克", "利福平0.15克", "盐酸二甲双胍", "盐酸"]} \end{CJK*}  &    {"Class": "Western Medicine", "Key Word": "Metformin", "Appearance": ["Sugar-coated or film-coated tablets, white after removal of coating"], "English Names": ["Yifupian", "Gehuazhi"], "Classification": ["Biguanide class", "Anti-tuberculosis drug"], "Specifications": ["0.25g"], "OTC Types": ["OTC Class B", "OTC Class A"], "Indications": ["Diabetes", "Obesity"], "Generic Name": ["Yifupian"], "Ingredients": ["Isoniazid and pyrazinamide", "Pyrazinamide", "0.1g pyrazinamide", "150mg pyrazinamide", "This product is a compound preparation", "Isoniazid", "300mg isoniazid", "0.15g isoniazid", "Metformin hydrochloride", "Hydrochloride"]}                                        \\ \hline
Symptom &   \{\begin{CJK*}{UTF8}{gbsn} {"中心词": "毛发脱落", "检查": ["毛发矿物质检查"], "相关疾病": ["斑秃", "慢性疲劳综合症"], "相关症状": ["毛发色淡而呈棕色", "毛发干燥易断", "皮肤变硬"], "所属科室": ["内科", "皮肤性病", "放疗、化疗科"], "发病部位": ["头部"]} \end{CJK*}      &  {"Key Word": "Hair Loss", "Examinations": ["Hair mineral analysis"], "Related Diseases": ["Alopecia areata", "Chronic Fatigue Syndrome"], "Related Symptoms": ["Hair color is light and brown", "Hair is dry and brittle", "Skin becomes hardened"], "Related Departments": ["Internal Medicine", "Dermatology and Venereology", "Radiation and Chemotherapy"], "Affected Area": ["Head"]} \\ \hline
\end{tabularx}
\caption{Knowledge cases in the CMeKG.}
\label{knowledge-label}
\end{table*}

\subsection{Knowledge-based Instruction Data}
Instruct-tuning has proven to be effective to tune large language models \cite{weifinetuned, ouyang2022training}, which helps the models perform satisfactorily under zero-shot scenarios with the cost of sufficient annotated instructions. Inspired by the automatic construction of the instruction along with the instances (inputs and outputs) \cite{wang2022self, alpaca}, we generate our instruction data based on the above medical knowledge. 

\begin{table}[h]
\renewcommand{\arraystretch}{1.2}
\centering
\begin{tabularx}{\columnwidth}{l}
\hline
\textbf{Instruction}: \cr Translate the following sentence into Chinese. \\ \hline
\textbf{Input}: \cr What are the possible reasons for liver cancer?      \\ \hline
\textbf{Output}: \cr \begin{CJK*}{UTF8}{gbsn}肝癌可能的原因有什么？  \end{CJK*}                                       \\ \hline
\end{tabularx}
\caption{Instance with an instruction.}
\label{instruction_input_output}
\end{table}

As demonstrated in Table \ref{instruction_input_output}, instruct-tuning involves supervised fine-tuning on the training instances and an instruction that describes the tasks in natural language. However, as for a large language model for medical dialogue, inputs are mostly stated as questions and instructions are all like ``Answer the following question''. Therefore, we dispose of the instructions and only preserve the inputs for our HuaTuo. 

\begin{table*}[htbp]
\begin{tabularx}{\textwidth}{l|X|X}
\hline
          & Instance (in Chinese) & Instances (translated to English)                                                                                                        \\ \hline
Question: & \begin{CJK*}{UTF8}{gbsn} 患者XXX因头部及眼后部疼痛并能听到连续不断的隆隆声来到医院，医生对其进行了检查后诊断为偏头痛。请问，该病有什么治疗方法？ \end{CJK*}                                                                         & Patient XXX came to the hospital due to pain in the head and behind the eyes, accompanied by continuous rumbling sounds. After an examination, the doctor diagnosed the patient with a migraine. What are the treatment options for this condition?     \\ \hline
Answer    & \begin{CJK*}{UTF8}{gbsn}治疗偏头痛的方法包括药物治疗和非药物治疗。药物治疗主要包括三大类：急性发作期治疗、预防性治疗和混合治疗。非药物治疗包括改变生活方式，如减少熬夜、避免过度劳累、避免饮酒、戒烟等；此外还包括针灸、按摩、脉冲放电治疗等。在治疗中应该综合评估患者的病情、临床表现、个人情况等因素，制定个性化的治疗方案。\end{CJK*} & The methods for treating migraines include medication and non-pharmacological therapies. Medication therapy primarily consists of three categories: acute attack treatment, preventative treatment, and mixed treatment. Non-pharmacological therapies include lifestyle changes such as reducing staying up late, avoiding overwork, avoiding alcohol consumption, and quitting smoking; in addition, they also include treatments such as acupuncture, massage, and pulse discharge therapy. During treatment, a comprehensive assessment of the patient's condition, clinical manifestations, personal circumstances, and other factors should be taken into account to develop a personalized treatment plan. \\ \hline
\end{tabularx}
\caption{Example for the supervised-fine-tuning training instances.}
\label{instruction_data}
\end{table*}

While the generated instructions are required to be diverse enough for unseen tasks \cite{wang2022self} in the general domain, the correctness of the fact in the responses from the large language model is of more concern in the biomedical domain \cite{gilson2023does}. Thus, we first sample knowledge instances from the knowledge graph and then generate the instances based on the specific knowledge with the OpenAI API \cite{chatgpt}. Finally, we collect over 8,000 instruction data, like examples in Table \ref{instruction_data} as training instances for supervised fine-tuning. 
\section{Experiment}
\subsection{Baselines}
In order to demonstrate the superior performance of HuaTuo, we conducted a comparative analysis with four baseline models.
\begin{itemize}
    \item \textbf{LLaMA} \cite{touvron2023llama} serves as the foundation model for our HuaTuo. In particular, we employed LLaMA-7B for its relative fairness in comparison to the other baselines and its ease of training.
    \item \textbf{Alpaca} \cite{alpaca} is an instructed-tuned version of LLaMA and boasts more than 80,000 instances generated in the general domain.
    \item \textbf{ChatGLM} \cite{zeng2023glm-130b} is an optimized dialogue model specifically designed for the Chinese chatting scenario. In our analysis, we compared HuaTuo's performance with ChatGLM-6B.
\end{itemize}

\subsection{Metrics}
For the generation tasks in the general domain, evaluation metrics, such as Bleu and Rouge are utilized to determine whether a generative model can produce responses similar to the ground truth. However, as for the medical QA tasks,  namely (1) safety, (2) usability, and (3) smoothness. Safety determines whether the response includes anything that can mislead the user into danger, such as wrong medicine recommendations. Usability reflects the medical expertise of a specific response. And, the Smoothness represents the ability as a language model.

In the domain of natural language generation, various evaluation metrics are utilized to assess the efficacy of generative models. The widely used metrics in the general domain include Bleu and Rouge, which compare generated responses with the ground truth. Additionally, for medical question-answering tasks, we introduce an evaluation metric, SUS. The SUS metric consists of three dimensions: \textbf{S}afety, \textbf{U}sability, and \textbf{S}moothness. The ``Safety'' dimension assesses whether the generated response has the potential to mislead the user and pose a danger to their health, for example, through incorrect medication recommendations. The ``Usability'' dimension evaluates the extent to which the generated response reflects medical expertise, while the ``Smoothness'' dimension measures the proficiency of the generative model as a language model.

\subsection{Results}

In this study, we constructed a test set of potential questions in Chinese dialogue scenarios and compared the generated responses of our HuaTuo model with three other baseline models.  To evaluate the model performance, we recruited five annotators with medical backgrounds who assessed the randomly mixed responses of the models using a three-point scale for each dimension of Safety, Usability, and Smoothness (SUS). The SUS scale ranges from 1 (not acceptable) to 3 (good), with 2 indicating an acceptable response. The average SUS scores are presented in Table \ref{evaluation}. Although LLaMA achieves the highest safety score, its responses are often uninformative and rephrased the questions, resulting in a low usability score. On the other hand, our HuaTuo model significantly improves knowledge usability without much compromising safety.

\begin{table*}[h]
\renewcommand{\arraystretch}{1.2}
\centering
\begin{tabular}{l|ccc}
\hline
& Safety & Usability & Smoothness \\ \hline
LLaMA  & \textbf{2.93}   & 1.21      & 1.58       \\ 
\hline
Alpaca & 2.64   & 2.05      & 2.30       \\ 
\hline
ChatGLM & 2.59   & 1.93      & 2.41       \\ 
\hline
\multicolumn{1}{l|}{HuaTuo (\begin{CJK*}{UTF8}{gbsn}华驼\end{CJK*})} 
& 2.88   & \textbf{2.12}      & \textbf{2.47}       \\ \hline
\end{tabular}

\caption{Experimental results of SUS score for the models. }
\label{evaluation}
\end{table*}

\section*{Ethics Statement}
The HuaTuo initiative is primarily dedicated to research and is not intended to offer medical advice. The medical information utilized in this study is sourced from an open-access medical knowledge graph. It is important to note that the accuracy of responses generated by large language models cannot be guaranteed, and the medical knowledge utilized therein should not be construed as a substitute for professional medical advice. If one experiences any discomfort or distress, it is strongly advised to seek the guidance of a qualified medical professional.

\bibliography{custom}

\begin{thebibliography}{13}
\expandafter\ifx\csname natexlab\endcsname\relax\def\natexlab#1{#1}\fi

\bibitem[{Chen et~al.(2023)Chen, Chen, Zhang, Jiang, Chen, Yu, Wang, Liang,
  Zhang, Zhang, Li, Wan, Li, and Wang}]{llm-zoo-2023}
Zhihong Chen, Junying Chen, Hongbo Zhang, Feng Jiang, Guiming Chen, Fei Yu,
  Tiannan Wang, Juhao Liang, Chen Zhang, Zhiyi Zhang, Jianquan Li, Xiang Wan,
  Haizhou Li, and Benyou Wang. 2023.
\newblock Llm zoo: democratizing chatgpt.
\newblock \url{https://github.com/FreedomIntelligence/LLMZoo}.

\bibitem[{Du et~al.(2022)Du, Qian, Liu, Ding, Qiu, Yang, and Tang}]{du2022glm}
Zhengxiao Du, Yujie Qian, Xiao Liu, Ming Ding, Jiezhong Qiu, Zhilin Yang, and
  Jie Tang. 2022.
\newblock Glm: General language model pretraining with autoregressive blank
  infilling.
\newblock In \emph{Proceedings of the 60th Annual Meeting of the Association
  for Computational Linguistics (Volume 1: Long Papers)}, pages 320--335.

\bibitem[{Gilson et~al.(2023)Gilson, Safranek, Huang, Socrates, Chi, Taylor,
  Chartash et~al.}]{gilson2023does}
Aidan Gilson, Conrad~W Safranek, Thomas Huang, Vimig Socrates, Ling Chi,
  Richard~Andrew Taylor, David Chartash, et~al. 2023.
\newblock How does chatgpt perform on the united states medical licensing
  examination? the implications of large language models for medical education
  and knowledge assessment.
\newblock \emph{JMIR Medical Education}, 9(1):e45312.

\bibitem[{Li et~al.(2023)Li, Li, Zhang, Dan, and
  Zhang}]{yunxiang2023chatdoctor}
Yunxiang Li, Zihan Li, Kai Zhang, Ruilong Dan, and You Zhang. 2023.
\newblock \href {http://arxiv.org/abs/2303.14070} {Chatdoctor: A medical chat
  model fine-tuned on llama model using medical domain knowledge}.

\bibitem[{Odmaa et~al.(2019)Odmaa, Yunfei, Zhifang, Damai, Baobao, Sujian, and
  Hongying}]{Odmaa}
BYAMBASUREN Odmaa, YANG Yunfei, SUI Zhifang, DAI Damai, CHANG Baobao,
  LI~Sujian, and ZAN Hongying. 2019.
\newblock Preliminary study on the construction of chinese medical knowledge
  graph.
\newblock \emph{Journal of Chinese Information Processing}, 33(10):1--7.

\bibitem[{OpenAI(2022)}]{chatgpt}
OpenAI. 2022.
\newblock Chatgpt.
\newblock \url{https://chat.openai.com}.

\bibitem[{Ouyang et~al.(2022)Ouyang, Wu, Jiang, Almeida, Wainwright, Mishkin,
  Zhang, Agarwal, Slama, Ray et~al.}]{ouyang2022training}
Long Ouyang, Jeffrey Wu, Xu~Jiang, Diogo Almeida, Carroll Wainwright, Pamela
  Mishkin, Chong Zhang, Sandhini Agarwal, Katarina Slama, Alex Ray, et~al.
  2022.
\newblock Training language models to follow instructions with human feedback.
\newblock \emph{Advances in Neural Information Processing Systems},
  35:27730--27744.

\bibitem[{Taori et~al.(2023)Taori, Gulrajani, Zhang, Dubois, Li, Guestrin,
  Liang, and Hashimoto}]{alpaca}
Rohan Taori, Ishaan Gulrajani, Tianyi Zhang, Yann Dubois, Xuechen Li, Carlos
  Guestrin, Percy Liang, and Tatsunori~B. Hashimoto. 2023.
\newblock Stanford alpaca: An instruction-following llama model.
\newblock \url{https://github.com/tatsu-lab/stanford_alpaca}.

\bibitem[{Touvron et~al.(2023)Touvron, Lavril, Izacard, Martinet, Lachaux,
  Lacroix, Rozi{\`e}re, Goyal, Hambro, Azhar et~al.}]{touvron2023llama}
Hugo Touvron, Thibaut Lavril, Gautier Izacard, Xavier Martinet, Marie-Anne
  Lachaux, Timoth{\'e}e Lacroix, Baptiste Rozi{\`e}re, Naman Goyal, Eric
  Hambro, Faisal Azhar, et~al. 2023.
\newblock Llama: Open and efficient foundation language models.
\newblock \emph{arXiv preprint arXiv:2302.13971}.

\bibitem[{Wang et~al.(2022)Wang, Kordi, Mishra, Liu, Smith, Khashabi, and
  Hajishirzi}]{wang2022self}
Yizhong Wang, Yeganeh Kordi, Swaroop Mishra, Alisa Liu, Noah~A Smith, Daniel
  Khashabi, and Hannaneh Hajishirzi. 2022.
\newblock Self-instruct: Aligning language model with self generated
  instructions.
\newblock \emph{arXiv preprint arXiv:2212.10560}.

\bibitem[{Wei et~al.(2022)Wei, Bosma, Zhao, Guu, Yu, Lester, Du, Dai, and
  Le}]{weifinetuned}
Jason Wei, Maarten Bosma, Vincent Zhao, Kelvin Guu, Adams~Wei Yu, Brian Lester,
  Nan Du, Andrew~M Dai, and Quoc~V Le. 2022.
\newblock Finetuned language models are zero-shot learners.
\newblock In \emph{International Conference on Learning Representations}.

\bibitem[{Xiong et~al.()Xiong, Wang, Zhu, Zhao, Liu, Wang, and
  Shen}]{xiong2023doctorglm}
Honglin Xiong, Sheng Wang, Yitao Zhu, Zihao Zhao, Yuxiao Liu, Qian Wang, and
  Dinggang Shen.
\newblock Doctorglm: Fine-tuning your chinese doctor is not a herculean task.

\bibitem[{Zeng et~al.(2023)Zeng, Liu, Du, Wang, Lai, Ding, Yang, Xu, Zheng,
  Xia, Tam, Ma, Xue, Zhai, Chen, Liu, Zhang, Dong, and Tang}]{zeng2023glm-130b}
Aohan Zeng, Xiao Liu, Zhengxiao Du, Zihan Wang, Hanyu Lai, Ming Ding, Zhuoyi
  Yang, Yifan Xu, Wendi Zheng, Xiao Xia, Weng~Lam Tam, Zixuan Ma, Yufei Xue,
  Jidong Zhai, Wenguang Chen, Zhiyuan Liu, Peng Zhang, Yuxiao Dong, and Jie
  Tang. 2023.
\newblock \href {https://openreview.net/forum?id=-Aw0rrrPUF} {{GLM}-130b: An
  open bilingual pre-trained model}.
\newblock In \emph{The Eleventh International Conference on Learning
  Representations (ICLR)}.

\end{thebibliography}
\bibliographystyle{acl_natbib}

\end{document}